\pdfoutput=1
\documentclass[11pt]{article}
\usepackage[utf8]{inputenc}
\usepackage[T1]{fontenc}
\usepackage[english]{babel}
\usepackage[margin=1in]{geometry}
\usepackage[numbers,sort&compress]{natbib}
\usepackage{amsmath,amssymb}
\usepackage{microtype}
\usepackage{graphicx}
\usepackage{subcaption}
\usepackage{booktabs}
\usepackage{multirow}
\usepackage{threeparttable}
\usepackage{rotating}
\usepackage{dblfloatfix}
\usepackage{environ}
\usepackage[hidelinks]{hyperref}
\usepackage{xcolor}
\usepackage{enumitem}
\usepackage{authblk}
\graphicspath{{figs/}}
\newsavebox{\AutoTableBox}
\NewEnviron{AutoTable}[3][]{%
  \sbox{\AutoTableBox}{\BODY}%
  \ifdim\wd\AutoTableBox>\columnwidth
    \begin{table*}[#1]\centering
      \caption{#2}\label{#3}
      \usebox{\AutoTableBox}
    \end{table*}%
  \else
    \begin{table}[#1]\centering
      \caption{#2}\label{#3}
      \usebox{\AutoTableBox}
    \end{table}%
  \fi
}

\newcommand{\ind}{\mathbf{1}}
\newcommand{\mAPfive}{\mathrm{mAP}\text{-}5}
\newcommand{\mAPfour}{\mathrm{mAP}\text{-}4}
\newcommand{\mAPthree}{\mathrm{mAP}\text{-}3}
\newcommand{\best}[1]{\textbf{#1}}

\title{\bfseries Coarse composition suffices: tabular in-context learning for multi-activity antimicrobial peptide profiling}

\author[1, $\dagger$]{Anuj Pal}
\author[1, $\dagger$]{Raunak Kumar}
\author[2, $\dagger$]{Dhruvi Solanki}
\author[3]{Parikshit Pareek}
\author[2,$\ast$]{Juhi Singh}
\author[4,$\ast$]{Jitin Singla}

\affil[1]{Department of Electronics and Communication Engineering, Indian Institute of Technology Roorkee, Roorkee 247667, Uttarakhand, India}
\affil[2]{Department of Bioengineering, Indian Institute of Science, Bengaluru 560012, Karnataka, India}
\affil[3]{Department of Electrical Engineering, Indian Institute of Technology Roorkee, Roorkee 247667, Uttarakhand, India}
\affil[4]{Department of Biosciences and Bioengineering, Indian Institute of Technology Roorkee, Roorkee 247667, Uttarakhand, India}
\affil[$\dagger$]{These authors contributed equally.}
\affil[$\ast$]{Corresponding authors.}

\date{}

\begin{document}
\maketitle

\begin{abstract}
\noindent
Antimicrobial peptides (AMPs) often act against multiple pathogen classes, making multi-label activity prediction a more realistic screening target than binary antimicrobial classification. The ESCAPE benchmark formalizes this setting, but leading approaches typically rely on multimodal, structure-conditioned deep models that are costly to train and tune. We show that a simple, sequence-only pipeline can match and surpass these methods by combining 330 interpretable sequence descriptors with TabPFN, a tabular foundation model that performs in-context prediction in a single forward pass without gradient-based training or hyperparameter search. On ESCAPE (82{,}359 peptides; five labels), a label-powerset TabPFN model achieves $\mAPfive=77.8\%$, improving on the previously best reported 72.1\%. A probabilistic classifier chain is the first method to match or exceed the best published average precision on each of the five labels simultaneously. The gains persist under the prior state-of-the-art single-fold training protocol, indicating they are not a training-set-size artefact, and are largest for remote homologues ($+11.2$ points below 30\% sequence identity). Ablations further show that predicted structure is unnecessary at inference and that performance is not driven by any single descriptor family: ten global physicochemical scalars recover 91\% of full-feature performance. Finally, explicitly modelling label dependence yields targeted benefits for scarce activities and supports ranking which activity to assay next from partial positive evidence.
\end{abstract}

\vspace{0.5em}
\noindent\textbf{Keywords:} antimicrobial peptides; multi-label classification;
tabular foundation models; benchmarking; sequence descriptors; assay prioritization

\section{Introduction}
\label{sec:intro}

Bacterial antimicrobial resistance was associated with an estimated 4.95 million deaths worldwide in 2019, of which 1.27 million were directly attributable \citep{murray2022}. Antimicrobial peptides (AMPs), typically cationic sequences of 10--50 residues, are among the most actively pursued alternatives to conventional antibiotics. Most AMPs act by physically disrupting microbial membranes rather than by engaging a single molecular target, a mechanism against which resistance is thought to evolve comparatively slowly \citep{zasloff2002,brogden2005,hancock2006}. A defining property of the class is that a single peptide is frequently active against more than one kind of pathogen: bacteria, fungi, viruses, and parasites \citep{yeaman2003}. The prediction target that matters for screening is therefore a multi-activity profile, not an isolated binary activity.

Computational AMP prediction has nonetheless been dominated by binary antimicrobial-versus-non-antimicrobial classification. Random forests over composition--transition--distribution descriptors \citep{bhadra2018,lawrence2021}, attention-augmented recurrent networks \citep{li2022amplify}, and fine-tuned protein language models \citep{lee2023ampbert} all address that task, and comparative assessments have repeatedly found them difficult to separate once evaluation is standardized \citep{xu2021assessment}. Multi-activity prediction is a more recent development. Graph neural networks over molecular representations \citep{ruizpuentes2022}, transformer architectures with asymmetric losses for label imbalance \citep{pang2022transimbamp}, and multi-task convolutional networks \citep{xu2023iampcn} predict several activities jointly. The ESCAPE benchmark \citep{escape2025} has recently standardized the task by integrating 82{,}359 peptides from 27 curated repositories under a fixed split and reporting a multimodal transformer baseline that utilizes both sequence and predicted three-dimensional structure. That benchmark is a substantial community resource: it fixes the split, retrains every prior method under a single protocol, and reports per-label as well as macro results. An informative feature of those results is that no method dominates, i.e., the leading method on macro average precision is not the leading method on any of antibacterial, antiviral, or antifungal activity, and different architectures lead on different labels.

The prevailing response to this task has been to add capacity and modalities. Structure-conditioned and language-model-conditioned architectures are now standard in peptide property prediction \citep{sun2025ssfgm,yu2024toxgin,lin2023}, and each addition carries real cost, i.e., structure prediction for every candidate, GPU training, architecture search, and hyperparameter tuning that must itself be validated. Recent critical assessments in adjacent domains have shown that apparently large deep-learning advantages can shrink or disappear once simple baselines, homology control, and evaluation protocol are examined \citep{bernett2024,sidorczuk2022,fernandezdiaz2024}, and the same examination has not been carried out for multi-activity AMP prediction.

In this study we ask how far a deliberately simple, sequence-only model can be taken on ESCAPE, and what that performance implies about the benchmark itself. We pair 330 interpretable sequence descriptors with TabPFN,a prior-data fitted network pre-trained on millions of synthetic tabular classification tasks that takes a labelled reference set as \emph{in-context samples} and returns calibrated posteriors in a single forward pass, with no gradient updates and no hyperparameter search on the target data \citep{hollmann2023,hollmann2025,mueller2022pfn}. We (i) evaluate standard multi-label transforms on a fixed descriptor matrix under the ESCAPE split and protocol; (ii) separate model quality from the quantity of labelled data available, by reproducing ESCAPE's released multimodal baseline and repeating every comparison under its single-fold protocol; (iii) stratify by sequence identity to quantify generalization to remote homologues; (iv) test whether performance is driven by any particular descriptor family and whether structural input contributes at inference; and (v) measure when modelling label dependence helps, including for prioritizing assays from partial positive evidence.

Overall, the results favour simplicity. A label-powerset TabPFN model improves five-label macro average precision to 77.8\% versus 72.1\% for the previously best-performing method, and a probabilistic classifier chain is the first approach to match or exceed the best published average precision on all five labels simultaneously. These gains persist under the matched single-fold protocol, are largest for peptides remote from the peptides supplied in context, are not attributable to any single descriptor family, and remain largely intact with only ten global physicochemical scalars (91\% of full-feature performance).

\section{Materials and methods}
\label{sec:methods}

\subsection{Dataset}
\label{sec:data}
We used the Expanded Standardized Collection for Antimicrobial Peptide Evaluation (ESCAPE) benchmark \citep{escape2025}. ESCAPE contains 82{,}359 peptides with four activity labels (antibacterial, antifungal, antiviral, antiparasitic) and a deterministic parent Antimicrobial label ($y_{\mathrm{AM}}=\bigvee_j y_j$). We used the predefined split: folds~1--2 (65{,}870 peptides) supply the in-context samples and the held-out fold ($n = 16{,}489$) is used only for testing. Label counts and prevalence in the test fold are summarized in Table~\ref{tab:composition}; antiparasitic is the rarest label and thus constrains macro-averaged performance (see Section~\ref{sec:sota}).

\begin{table}
\centering
\small
\caption{Label structure of the ESCAPE benchmark. Positive counts and prevalence are
given for the held-out test fold ($n = 16{,}489$); the conditional column is the
fraction of antimicrobial test peptides carrying each activity.}
\label{tab:composition}
\begin{tabular}{lrrr}
\toprule
Label & Test positives & Prevalence (\%) & $P(\text{label} \mid \text{AM})$ (\%)\\
\midrule
Antibacterial  & 3{,}187 & 19.33 & 75.3\\
Antifungal     & 1{,}316 &  7.98 & 31.1\\
Antiviral      &   946   &  5.74 & 22.3\\
Antiparasitic  &    77   &  0.47 &  1.8\\
Antimicrobial & 4{,}235 & 25.68 & 100.0 \\
\bottomrule
\end{tabular}
\end{table}

\subsection{Evaluation metrics and statistics}
\label{sec:metrics}
Three averaging scales are used and are not interchangeable. We report $\mAPfive$ (all five labels) to enable direct comparison against published results. For dependence analyses we use $\mAPfour$, which averages over the four real activities and excludes the deterministic Antimicrobial parent. We also report $\mAPthree$, which averages over antibacterial, antifungal, and antiviral only (excluding both the parent and Antiparasitic), because Antiparasitic has only 77 test positives. Unless stated otherwise, every average precision, macro average, per-label difference, bootstrap interval, and threshold-dependent score is reported on the $0$--$100$ scale.

Interval estimates use 2{,}000 bootstrap resamples of the test peptides. All transforms are scored on \emph{identical} resamples, and within each resample the macro metric is averaged across the three seeds, treating seeds as repeated measurements of one test set rather than as additional independent data. A difference is called significant when its two-sided 95\% percentile interval excludes zero. Point estimates are means over three seeds and $\pm$ denotes the sample standard deviation.

\subsection{Feature representation}
\label{sec:features}
Each peptide was encoded as a 330-dimensional numeric vector partitioned into eight interpretable families (Table~\ref{tab:features}). Feature extraction was identical for every transform compared here, so no difference between transforms can arise from representation. Composition--transition--distribution (CTD) descriptors encode three complementary views of a physicochemical property along the sequence: \emph{composition} is the fraction of residues in each property class, \emph{transition} the frequency of switches between classes at adjacent positions, and \emph{distribution} the sequence positions at which the first, 25\%, 50\%, 75\%, and 100\% of a class are reached. A descriptor such as \texttt{Hydrophobicity\_D\_3\_0} is therefore spatial rather than compositional.

\begin{table}
\centering
\small
\caption{Feature representation (330 numeric columns).}
\label{tab:features}
\begin{tabular}{llr}
\toprule
Family & Content & $n$ \\
\midrule
ESM      & ESM-2 embeddings, mean-pooled and PCA-reduced \citep{lin2023} & 128 \\
CTD      & 7 properties $\times$ 21 (composition 3, transition 3, distribution 15) & 147 \\
AAC      & Amino-acid composition & 20 \\
AAIndex  & PCA-reduced AAindex residue-property matrices \citep{kawashima2008} & 15 \\
PhysChem & Length, net charge, charge density, p$I$, instability index, aromaticity, & 10 \\
         & aliphatic index, Boman index, hydrophobic ratio, hydrophobic moment & \\
Motif    & R-X-R, RW repeat, RGD, C-X$_3$-C, RKKR & 5 \\
Density  & Arg, Trp, Cys, Pro residue densities & 4 \\
Cleavage$^{a}$ & Basic-cleavage-pair count & 1 \\
\midrule
\multicolumn{2}{l}{\textbf{Total}} & \textbf{330} \\
\bottomrule
\end{tabular}
\begin{minipage}{0.72\textwidth}
\vspace{2pt}
\footnotesize
$^{a}$The basic-cleavage-pair count is held out as its own single-column family
rather than folded into PhysChem
\end{minipage}
\end{table}

\subsection{Backbone}
\label{sec:backbone}
TabPFN is a prior-data fitted network: a transformer pre-trained on millions of synthetic tabular classification tasks that receives an entire labelled reference set as in-context samples and returns test-row posteriors in a single forward pass, with no gradient updates and no hyperparameter search on the target data \citep{hollmann2023,hollmann2025}. Because such networks approximate the Bayesian posterior predictive distribution \citep{mueller2022pfn}, their probability outputs are calibrated by construction, which is what licenses the conditional analyses in Sections~\ref{sec:dependence-methods} and~\ref{sec:pu-methods}.

Two terms, \emph{fit} and \emph{inference}, are used in a TabPFN-specific sense. A \emph{fit} performs no weight updates, it only loads the labelled reference peptides into the model context, so the ``fitted'' object is the stored context matrix plus fixed pre-trained weights. \emph{Inference} is a single forward pass that reads a test peptide with that context and returns a posterior over labels. We therefore treat the 65{,}870 peptides of folds~1--2 as \emph{in-context samples}, and reserve \emph{training} for gradient-based procedures (TabPFN's one-off synthetic pre-training and the deep-learning baselines). No gradients are computed on peptide data. Reported fit time is context ingestion (not optimisation), hence typically much smaller than inference time.

We used TabPFN v3 (\texttt{tabpfn} 8.2.0) with $n_{\mathrm{estimators}} = 16$ and seeds $\{42, 1665, 8914\}$, matching the seed protocol of the published baseline. Ensemble classifier chains use $n_{\mathrm{estimators}} = 8$ to make eight chains affordable. No hyperparameter was tuned on ESCAPE at any stage. Experiments ran on two NVIDIA RTX PRO 5000 Blackwell GPUs (48\,GB each) under PyTorch 2.13.0+cu130, scikit-learn 1.7.2, NumPy 2.2.6, and Python 3.10.20.

\subsection{Multi-label transforms}
\label{sec:strategies}
Five standard transforms were compared on an identical feature matrix, split, and backbone, differing only in how the multi-label target is factorized. A worked out example for the transforms is given in Supplementary Section~\ref{sec:s1}.

\paragraph{Notation.} Let $\mathbf{x} \in \mathbb{R}^{330}$ denote a peptide's descriptor vector and $\mathbf{y} \in \{0,1\}^{L}$ its label vector (with $L = 5$ for the benchmark comparison and $L = 4$ in the dependence analyses). Labels are ordered frequency-descending, $y_1 = $ Antibacterial, $y_2 = $ Antifungal, $y_3 = $ Antiviral, $y_4 = $ Antiparasitic, $y_5 = $ Antimicrobial. $\hat{P}$ denotes a TabPFN posterior probability.

\textbf{Binary relevance (BR)} assumes the labels are conditionally independent given the features and factorizes the joint as a product of marginals, \begin{equation} \hat{P}_{\mathrm{BR}}(\mathbf{y} \mid \mathbf{x}) = \prod_{j=1}^{L} \hat{P}(y_j \mid \mathbf{x}), \label{eq:br} \end{equation} fitting one classifier per label. It encodes no dependence and is the conditional-independence floor against which the others are measured. BR costs $L$ inference passes.

\textbf{Label powerset (LP)} treats a peptide's whole activity profile as one category rather than as five separate questions \citep{tsoumakas2007}. A peptide active against bacteria and fungi but not against viruses or parasites has the profile $(1,1,0,0,1)$, the last entry being Antimicrobial, which is 1 whenever any of the four activities is. Because that parent is fixed by the other four, only $2^4 = 16$ of the $2^5$ conceivable profiles can occur, from all-negative to active against all four. LP therefore solves a single 16-class problem and returns one probability per profile. A per-label probability is recovered by adding up the profiles that contain the label of interest: \begin{equation} \hat{P}_{\mathrm{LP}}(y_j = 1 \mid \mathbf{x}) = \sum_{c \in \mathcal{C} : c_j = 1} \hat{P}(c \mid \mathbf{x}), \label{eq:lp} \end{equation} where $\mathcal{C}$ is the set of profiles seen in context and $c_j$ the $j$th entry of profile $c$, so the sum runs over every profile in which label $j$ is present. LP can represent any pattern of co-occurrence among the profiles it has seen, at the cost of assigning zero probability to a profile absent from context. LP only cost one inference pass.

\textbf{Probabilistic classifier chains (PCC)} predict the labels in a fixed order, each one conditioned on the answers already given, i.e., is the peptide antibacterial, then, given that answer, is it antifungal, and so on. Multiplying these conditionals gives the probability of one complete profile, and PCC evaluates all $2^5 = 32$ profiles rather than following a single path \citep{dembczynski2010}: \begin{equation} \hat{P}_{\mathrm{PCC}}(\mathbf{y} \mid \mathbf{x}) = \prod_{j=1}^{L} \hat{P}\!\left(y_j \mid \mathbf{x}, y_1, \dots, y_{j-1}\right), \qquad \hat{P}(y_j = 1 \mid \mathbf{x}) = \!\!\sum_{\mathbf{y}: y_j = 1}\!\! \hat{P}(\mathbf{y} \mid \mathbf{x}). \label{eq:pcc} \end{equation} As in LP, a per-label probability is a sum over the profiles containing that label. Enumerating 32 profiles makes PCC the most expensive transform here, but it yields the full joint posterior the conditional analyses require. Unlike LP it is not confined to the 16 profiles the Antimicrobial rule permits, which is why it can return incoherent ones (Section~\ref{sec:dependence}). Cost:
$L$ fits, $2^L$ inference passes, i.e., 4{,}116\,s against 128\,s for LP.
 
\textbf{Classifier chains (CC)} keep the same label order as PCC, but they follow only one trajectory through the chain. After predicting $y_1$, the model feeds that prediction into the classifier for $y_2$, then feeds the prediction for $y_2$ into the classifier for $y_3$, and so on \citep{read2011}. This avoids enumerating all profiles, making CC much cheaper than PCC, but it also means mistakes made early can cascade downstream.

Formally,
\begin{equation}
\hat{P}_{\mathrm{CC}}(\mathbf{y} \mid \mathbf{x}) = \prod_j \hat{P}(y_j \mid \mathbf{x}, y_{<j}),
\end{equation}
where $y_{<j}$ are the earlier labels, given as ground truth in the in-context samples and replaced by the chain's own predictions at inference. How the earlier predictions are encoded is therefore important. \emph{Hard} context appends the binarized indicator $\ind[\hat{p}_{<j} > \tau] \in \{0,1\}$, matching the 0/1 encoding of the in-context labels, while \emph{soft} context appends the raw probability $\hat{p}_{<j} \in [0,1]$.
 
\textbf{Ensemble classifier chains (ECC)} average the per-label probabilities of eight chains under independently sampled label orders, so that no single arbitrary ordering decides the result. ECC run at $n_{\mathrm{estimators}} = 8$ to make eight chains affordable.

\subsection{ESCAPE baseline re-run and matched-protocol evaluation}
\label{sec:repro-methods}
Published per-label numbers for the ESCAPE baseline \citep{escape2025} are point estimates from the original authors' runs, so comparisons to them are unpaired. For a paired comparison, we re-ran the released ESCAPE baseline checkpoints (\texttt{Best\_model\_Fold1.pth}, \texttt{Best\_model\_Fold2.pth}) on our hardware without any retraining. Structural maps are available for 5{,}495 of the 16{,}489 test peptides (33.3\%), so paired analyses are restricted to this subset.

\paragraph{Matched-protocol evaluation:} The ESCAPE baseline averages two checkpoints trained on opposite folds, whereas our transforms take both folds as in-context samples. To match protocols, we supplied each fold separately as context (per seed) and averaged the two resulting probability matrices, mirroring the baseline. The gap to the full-context experiment isolates the effect of the additional in-context samples.

\paragraph{Structural-branch ablation:} To test whether the ESCAPE baseline's structural modality contributes at inference, we held the sequence input fixed and replaced only the structural input with (i) an all-zero image, (ii) uniform noise, and (iii) another peptide's real map, which preserves input statistics while destroying the association between a peptide and its own structure. Changes were quantified both as per-label average precision and as the maximum and mean absolute change in predicted probability over every peptide~$\times$~label pair, against the same checkpoint's real-map run.

\subsection{Quantifying label dependence}
\label{sec:dependence-methods}
Two analyses establish whether conditional dependence between activities exists and what it is worth. Both are restricted to the four real activities, excluding the deterministic Antimicrobial parent, whose inclusion inflates apparent dependence two-to-four fold.
 
\emph{An oracle conditional-predictability ceiling} scores each activity with its three true sibling labels appended to the features, and compares that against features alone. This upper-bounds the gain available to any method that exploits sibling labels. Recovery is reported as $(\text{transform} - \text{BR}) / (\text{oracle} - \text{BR})$ on the $\mAPfour$ scale. This measures what dependence is worth for \emph{per-label ranking}.
 
\emph{A joint-versus-product held-out likelihood} comparison asks the complementary question, whether dependence exists at all independently of any ranking metric. We compare the mean held-out log-likelihood of the true four-activity test vectors under the BR product of marginals against the PCC joint, computed in log space with log-sum-exp normalization, in nats per peptide. Log-likelihood is a strictly proper scoring rule, so a joint-model advantage on it constitutes exploitable dependence by construction, whether or not that dependence is recoverable as average precision.

\subsection{Homology stratification}
\label{sec:homology-methods}
Test peptides were searched against the in-context set (folds~1--2) with MMseqs2 \citep{steinegger2017} 18.8cc5c \texttt{easy-search} (\texttt{-s 7.5 -e 10000 --max-seqs 4000 --alignment-mode 3}) and each test peptide labelled with the maximum sequence identity among hits with query coverage $\geq 0.5$. Peptides with no qualifying hit were assigned zero. There are no exactly duplicated sequences within or across splits.

\subsection{Feature-group attribution}
\label{sec:attribution-methods}
Three analyses answer different questions and are expected to disagree where features are redundant. \emph{Group-only} refits on each family alone and measures standalone sufficiency. \emph{Leave-one-group-out} refits without each family and measures what is lost when it is unavailable from the start. \emph{Permutation} shuffles the family's columns in the test matrix and re-infers, measuring the model's reliance on it. All arms run on a stratified 4{,}000-peptide test subsample drawn once and shared across three analyses, retaining all 77 antiparasitic positives. Label prevalence in the subsample is consequently higher than in the full test set, so its absolute values are not comparable to Section~\ref{sec:sota} and are used only for within-experiment contrasts.

\subsection{Positive-unlabelled assay prioritization}
\label{sec:pu-methods}
Curated activity repositories record what was found positive. An activity a peptide was never assayed for is unlabelled rather than known-negative \citep{elkan2008,bekker2020}. We therefore treated confirmed positives as reliable and all zeros as unlabelled, and scored the task a screening campaign actually faces: given a peptide already confirmed positive for $m$ activities, rank the remaining activities by which to assay next.

For each test peptide with at least $m$ confirmed positives and at least one remaining candidate, the $m$ highest-prevalence positives were revealed and every unobserved activity scored by the joint posterior $\hat{P}(a = 1 \mid \mathbf{x}, \text{observed positives})$. Rankers compared were BR (which cannot use the revealed positives and is therefore the do-nothing baseline) and the LP and PCC joint posteriors. Performance is reported as Hit@1, mean reciprocal rank, and a pooled positive-unlabelled AP, which is a lower bound because the unlabelled set contains an unknown number of true positives. As a negative control the conditioning was permuted across peptides, supplying the same quantity of label information while severing its link to the specific peptide. This control is a \emph{single} donor assignment per $m$, reused across seeds, not a distribution over permutations. It is therefore a point comparison and yields no permutation $p$-value. Hit@1 is additionally reported decomposed by which activity is held out, because the pooled figure is dominated by the commonest target.

\section{Results}
\label{sec:results}

\subsection{Improved average precision across all activity labels}
\label{sec:sota}
\begin{figure*}
\centering
\includegraphics[width=\textwidth]{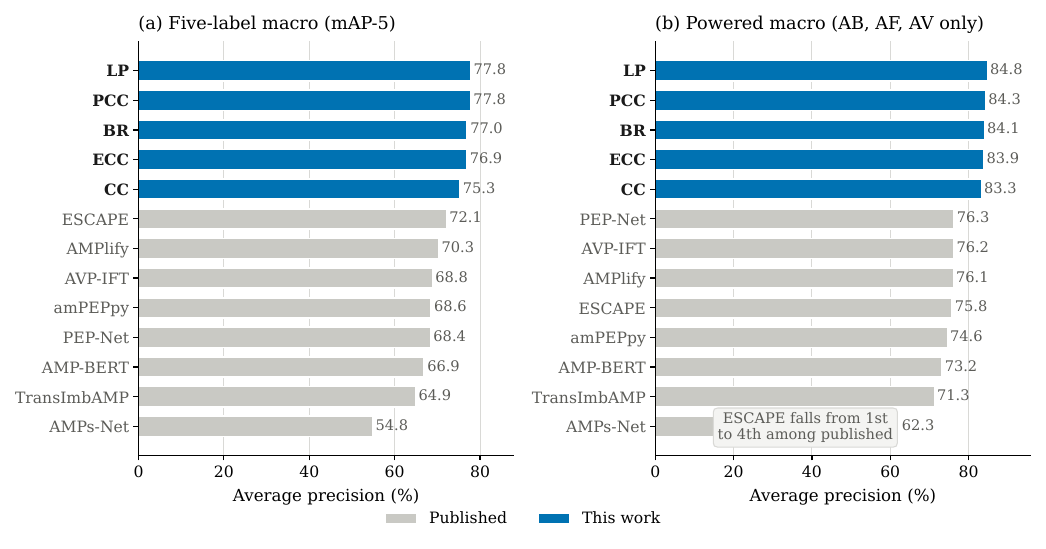}
\caption{Each transform leads the published benchmarks on both macro scales. (a) Five-label macro average precision, (b) The macro restricted to the three adequately powered activities (antibacterial, antifungal, antiviral). Values are recomputed from the per-label average precisions in Table~\ref{tab:frontier}.}
\label{fig:frontier}
\end{figure*}

Table~\ref{tab:frontier} places the five transforms against every published result on the ESCAPE benchmark. The label powerset transform reaches a five-label macro average precision of $77.79 \pm 0.23$ and the probabilistic classifier chain $77.76 \pm 0.12$, against 72.12 for the previously best-performing method, a margin of 5.67 and 5.64 percentage points respectively. Binary relevance, which encodes no dependence between activities at all, reaches $77.01 \pm 0.17$ and is already ahead of every published method (Figure~\ref{fig:frontier}a). All five transforms use the same 65{,}870 peptides as in-context samples and involve no gradient training (Section~\ref{sec:backbone}). Supplementary Section~\ref{sec:s4} records what each published method's own publication set out to predict, since several were binary or single-activity, classifiers were adapted to the five-label output by ESCAPE's benchmark authors.

\begin{table*}
\centering
\small
\caption{Average precision (\%) on the ESCAPE test set ($n = 16{,}489$), averaged over seeds $\{42, 1665, 8914\}$. Per-label published values are as reported by the benchmark authors. Bold marks the best value in each column. \emph{Powered macro} averages Antibacterial, Antifungal, and Antiviral only. $\mAPfive$ and \emph{Powered} are recomputed here from the per-label columns so that every row is on the same footing. Per-seed values in Supplementary Section~\ref{sec:s2}.}
\label{tab:frontier}
\begin{tabular}{lrrrrrrr}
\toprule
Method & Antibact. & Antifung. & Antiviral & Antiparas. & Antimicrob. & $\mAPfive$ & Powered \\
\midrule
\multicolumn{8}{l}{\emph{Published}} \\
AMPs-Net \citep{ruizpuentes2022}       & 82.5 & 53.1 & 51.2 & 5.3  & 82.1 & 54.8 & 62.3 \\
TransImbAMP \citep{pang2022transimbamp}& 92.5 & 56.3 & 65.0 & 16.7 & 94.0 & 64.9 & 71.3 \\
AMP-BERT \citep{lee2023ampbert}        & 92.3 & 61.5 & 65.9 & 21.4 & 93.6 & 66.9 & 73.2 \\
PEP-Net \citep{han2024pepnet}          & 95.2 & 72.6 & 61.2 & 16.2 & 96.7 & 68.4 & 76.3 \\
amPEPpy \citep{lawrence2021}           & 93.9 & 62.2 & 67.7 & 23.8 & 95.2 & 68.6 & 74.6 \\
AVP-IFT \citep{guan2024avpift}         & 94.3 & 63.3 & 71.1 & 20.0 & 95.5 & 68.8 & 76.2 \\
AMPlify \citep{li2022amplify}          & 94.0 & 68.3 & 66.1 & 27.7 & 95.3 & 70.3 & 76.1 \\
ESCAPE \citep{escape2025}              & 94.2 & 63.4 & 69.8 & \best{37.6} & 95.6 & 72.12 & 75.8 \\
\midrule
\multicolumn{8}{l}{\emph{This work}} \\
CC   & 97.27 & 73.69 & 78.93 & 28.93 & 97.53 & 75.27 & 83.30 \\
ECC  & 97.21 & 74.73 & 79.85 & 34.84 & 97.81 & 76.89 & 83.93 \\
BR   & 97.27 & 74.98 & 79.95 & 34.76 & 98.10 & 77.01 & 84.07 \\
PCC  & 97.27 & 75.37 & 80.41 & \best{37.68} & 98.06 & 77.76 & 84.35 \\
LP   & \best{97.51} & \best{75.65} & \best{81.29} & 36.41 & \best{98.11} & \best{77.79} & \best{84.82} \\
\bottomrule
\end{tabular}
\end{table*}

\begin{figure*}
\centering
\includegraphics[width=\textwidth]{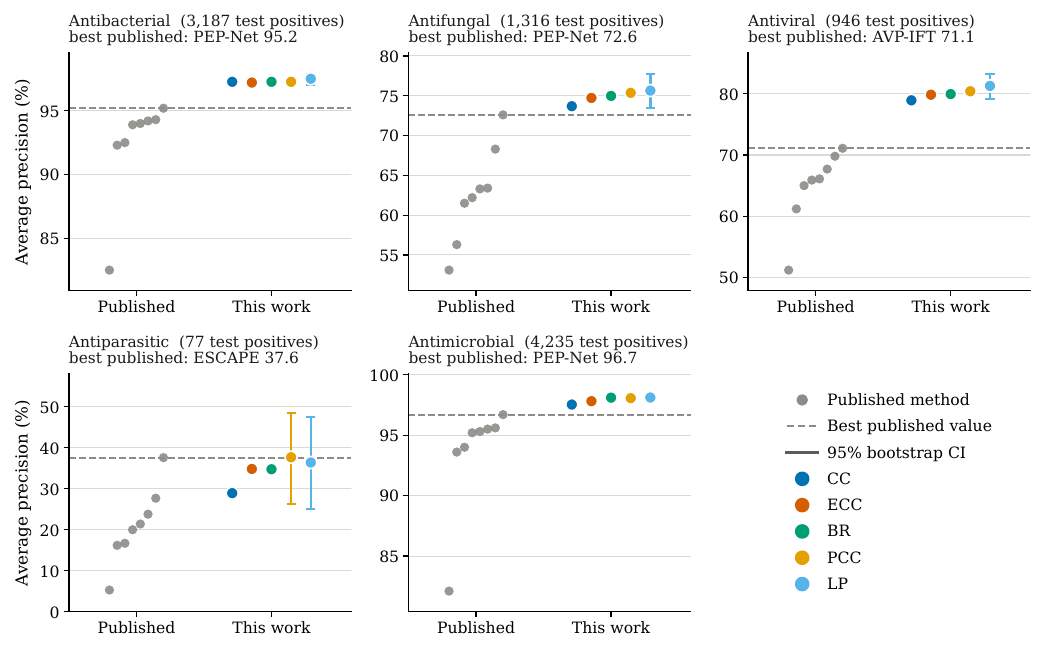}
\caption{Per-label average precision, all eight published methods against all five proposed transforms. Each panel is one activity label, annotated with its test-positive count. The dashed line is the best published value for that label. Every transform exceeds the best published value on antibacterial, antifungal, antiviral and antimicrobial activity. Error bars are 95\% bootstrap percentile intervals, shown for LP in every panel and additionally for PCC on Antiparasitic. They are narrower than the plotting symbol on Antibacterial and Antimicrobial. Note the differing $y$-axis range in each panel.}
\label{fig:perlabel}
\end{figure*}

The per-label picture is more informative than the macro. Among published methods no single architecture leads on more than three labels. PEP-Net leads on Antibacterial, Antifungal, and Antimicrobial, AVP-IFT on Antiviral, and ESCAPE on Antiparasitic only, despite ranking first overall. Every transform reported here exceeds the best published value on Antibacterial, Antifungal, Antiviral, and Antimicrobial, and PCC additionally matches it on Antiparasitic ($37.68$ against $37.6$). PCC is therefore the first model to match or exceed the best published average precision on all five labels simultaneously. The Antiparasitic comparison should be read with its power in mind: with 77 test positives, PCC's 95\% interval is $[26.26, 48.50]$, so the correct statement is parity rather than superiority. Because published per-label values are point estimates from the original authors' runs, they support only the question of whether each falls inside our 95\% confidence interval. On Antibacterial, Antifungal, Antiviral, and Antimicrobial, every published value lies strictly below the lower bound of every transform's interval. On Antiparasitic all published values lie inside our intervals for all five transforms (Figure~\ref{fig:perlabel}). Building on the per-label results, we focus on the “powered” macro (averaging only adequately powered labels) because Antiparasitic is so rare (77 positives) that its wide uncertainty can dominate the equal-weight five-label metric (Figure~\ref{fig:frontier}b).

These results were obtained by forward passes through the pre-trained TabPFN network alone, with no gradient training and no hyperparameter search. The labelled peptides supplied as in-context samples are read in the same forward pass as the test query and directly yield its predicted labels. Because the transforms differ only in how the label is presented, their costs differ by orders of magnitude: LP is both the most accurate transform and the cheapest, $5.0$ times cheaper at inference than BR, the next-cheapest, and $32$ times cheaper than PCC (Table~\ref{tab:cost}).

\begin{table}
\centering
\small
\caption{Computational cost of each transform on the full benchmark, single GPU, $n_{\mathrm{estimators}} = 16$ (8 for ECC). Times exclude feature extraction, which is shared across all transforms. A \emph{fit} is the ingestion of the 65{,}870 in-context samples and involves no gradient update; an \emph{inference pass} is the forward pass that returns posteriors for the test peptides (Section~\ref{sec:backbone}). Counts are given as a function of the number of labels $L = 5$. Times are means over seeds $\{42, 1665, 8914\}$}
\label{tab:cost}
\begin{tabular}{lrrrrr}
\toprule
Transform & Fits & Inference passes & Fit (s) & Inference (s) & $\mAPfive$ \\
\midrule
LP  & 1 & 1     & \best{18} & \best{128}   & \best{77.79} \\
BR  & $L$ & $L$ & 85 & 646   & 77.01 \\
CC  & $L$ & $L$ & 124 & 907 & 75.27 \\
ECC & $8L$ & $8L$ & 255 & 3{,}791 & 76.89 \\
PCC & $L$ & $2^L$ & 82 & 4{,}116 & 77.76 \\
\bottomrule
\end{tabular}
\end{table}

\subsection{Separating Data-Size Effects from Model Gains}
\label{sec:matched}
Our transforms take both non-test folds (65{,}870 peptides) as in-context samples, whereas ESCAPE, the previous state of the art, trains on one fold (${\approx}32{,}900$ peptides) at a time and averages the test metric. A margin measured that way therefore confounds model quality with twice as much labelled data. One observation motivated testing this directly: halving the number of in-context samples (a random half of folds~1--2) costs $5.01$ points of $\mAPfive$ for BR (4.71 for LP, 4.43 for CC; Supplementary Section~\ref{sec:s3}), the same magnitude as BR's full-context margin of $+5.57$ over the reproduced baseline. That subset was drawn at random across both folds, however, and is therefore not like-for-like with the baseline's single-fold protocol, which the fold-matched experiment below supplies.

Running the released ESCAPE checkpoints over the 5{,}495 test peptides for which structural maps were released gives macro AP $72.18$ against the published $72.12$, a difference of $0.06$ points, with every per-label value within $1.7$ points of its published counterpart. The released checkpoints and evaluation code therefore reproduce the publication closely, and the comparison below (Table~\ref{tab:matched}) is against a faithfully reproduced baseline.

\begin{table}
\centering
\small
\caption{Paired comparison against ESCAPE on the 5{,}495 test peptides with released structural maps, 2{,}000 shared bootstrap resamples. \emph{Fold-matched} supplies one fold at a time as in-context samples and averages the two probability matrices, mirroring the ESCAPE's own protocol. All entries are differences in $\mAPfive$, in percentage points.}
\label{tab:matched}
\begin{tabular}{lrrrl}
\toprule
Transform & $\Delta$ full-data & $\Delta$ fold-matched & 95\% CI (matched) & Significant \\
\midrule
LP  & $+6.08$ & $+3.69$ & $[+1.04, +6.46]$ & yes \\
BR  & $+5.57$ & $+2.86$ & $[+0.04, +5.87]$ & yes (marginal) \\
CC  & $+3.32$ & $-0.13$ & $[-3.21, +3.60]$ & no \\
PCC & $+6.01$ & $+3.26$ & $[+0.65, +6.03]$ & yes \\
\bottomrule
\end{tabular}
\end{table}

Under the matched protocol LP retains a significant advantage of $+3.69$ $[+1.04, +6.46]$ points, which also holds on $\mAPfour$ ($+4.09$ $[+0.81, +7.53]$). BR retains $+2.86$, significant on $\mAPfive$ but marginally so, and not significant on $\mAPfour$. CC is indistinguishable from the baseline on the macro. PCC retains $+3.26$ $[+0.65, +6.03]$ on $\mAPfive$ and $+3.56$ $[+0.29, +7.00]$ on $\mAPfour$. Direct paired measurement of the confound (full-data minus fold-matched, on identical test rows, $n = 16{,}489$) gives $+2.40$ $[+1.80, +3.02]$ for LP, $+3.08$ $[+2.16, +3.85]$ for BR, $+2.96$ $[+1.89, +3.88]$ for CC, and $+3.05$ $[+2.15, +3.85]$ for PCC; all four intervals exclude zero. Roughly half of the full-data margin is therefore attributable to the number of in-context samples, and the remainder to the model.

One secondary observation bears on architecture choice. The baseline gains $+5.86$ points of $\mAPfive$ from averaging its two checkpoints, more than twice what any of our transforms gain from the same operation (LP $+2.32$, BR $+2.27$, CC $+1.61$), and its individual checkpoints are weak in isolation (65.03 and 67.62 against 72.18 ensembled). For an in-context learner the reverse holds: one model on 65{,}870 rows outperforms two on ${\approx}32{,}900$ each (LP 77.79 against 75.40), which is the expected behaviour when additional labelled rows are additional evidence at inference rather than additional gradient steps.

\subsection{Remote-homologue peptides show strongest gains}
\label{sec:homology}
A margin obtained on a benchmark in which a third of the test set is nearly identical to sequences the model has seen in context invites the objection that it reflects memorization. The predefined fold split does place 5{,}342 test peptides (32.4\%) out of 16{,}489 at $\geq 90\%$ identity to an in-context sequence.

Paired per-peptide comparison of 5{,}495 test peptides against the previous state of the art, by identity band, shows the margin \emph{growing} as similarity falls. Below 30\% identity it is $+9.65$ points (BR), $+11.23$ (LP), and $+10.03$ (PCC), against $+3.8$ (LP) and several non-significant values in the $\geq 90\%$ band (Table~\ref{tab:strata}). Pooled over the whole covered subset the paired margins are $+6.08$ $[+3.13, +9.10]$ for LP, $+6.01$ $[+3.78, +8.15]$ for PCC, and $+5.57$ $[+3.15, +7.86]$ for BR, so the remote-homologue margin is close to twice the pooled one. The advantage is therefore largest exactly where the benchmark is hardest and where a screening campaign would actually operate.

\begin{table}
\centering
\small
\caption{Homology stratification of the predefined test split. \emph{Margin} is the
paired per-peptide difference in $\mAPfive$ against the previous state of the art,
in percentage points, restricted within each band to labels with at least five
positives in that band.}
\label{tab:strata}
\begin{tabular}{lrrrr}
\toprule
Identity band & Test peptides & \multicolumn{3}{c}{Margin over previous SOTA (points)} \\
\cmidrule(l){3-5}
              & $n$ (\% of test) & BR & LP & PCC \\
\midrule
$<30\%$      & 3{,}708 (22.5\%) & $+9.65$ & $+11.23$ & $+10.03$ \\
$\geq 90\%$  & 5{,}342 (32.4\%) & n.s. & $+3.8$ & n.s. \\
\midrule
All (paired) & 5{,}495 (100\%) & $+5.57$ & $+6.08$ & $+6.01$ \\
\bottomrule
\end{tabular}
\end{table}

\subsection{Sequence-only information is sufficient}
\label{sec:attribution}
To determine which parts of our $330$-dimensional descriptor set actually carry predictive signal (and thus whether the gains come from rich representations or from coarse composition), we ran an ablation-style attribution study on a stratified 4{,}000-peptide random subsample retaining all 77 antiparasitic positives (baseline $\mAPfive$ $80.41$ on this subsample; its absolute values are not comparable to Section~\ref{sec:sota}). Because rerunning TabPFN is computationally expensive, we restrict these refits to this smaller test subset. Table~\ref{tab:attribution} summarizes the three attribution experiments (group-only refits, leave-one-family-out refits, and test-time permutation within each family).

Refitting on each descriptor family alone shows that the signal is broadly distributed and largely redundant. Ten global physicochemical scalars reach $73.52$, recovering 91.4\% of full-feature performance; 20 amino-acid-composition features reach $77.06$ (95.8\%); 15 AAindex components reach $76.05$ (94.6\%); the 147 CTD descriptors reach $77.30$ (96.1\%); and the 128-dimensional ESM-2 protein language model block \citep{lin2023} reaches $79.19$ (98.5\%), within $1.22$ points of the full model. A learned embedding and a hand-crafted descriptor set are thus near-substitutable here. Only the three smallest families fall away: residue densities reach $51.12$ and the motif and cleavage features, which are near-binary indicators, reach $16.26$ and $16.48$.

Leave-one-family-out shows that no single feature family is essential. Removing ESM gives the biggest drop ($-0.74$ points), followed by CTD ($-0.67$) and every other family changes $\mAPfive$ by at most $0.21$. Dropping motif ($+0.51$) or cleavage ($+0.12$) even gives a tiny \emph{increase}. Permutation importance, however, makes CTD and ESM look much more important ($-45.19$ and $-19.17$ points). This happens because permuting a large, highly correlated block creates unrealistic (out-of-distribution) inputs, so the score drop can be exaggerated. For CTD, the permutation and refit results differ by a factor of 67, so we trust the refit results more. The small gains from removing motif and cleavage are within noise for a 4{,}000-peptide subsample, and should not be taken as evidence that these features are harmful.

This redundancy is the most direct explanation of why a simple model suffices on this benchmark. The accessible signal is coarse composition, and it is close to saturated by the time a few dozen descriptors have been supplied.

The same conclusion is reached independently from the released multimodal checkpoints. Holding the sequence input fixed and replacing only the structural input with an all-zero image, with uniform noise, or with another peptide's map moves ensemble macro AP by at most $0.056$ points across all 5{,}495 covered peptides. Essentially, all of that comes from Antiparasitic, whose own per-label spread is $0.298$ points and which therefore contributes up to $0.060$ points to the macro. The other four labels contribute at most $0.005$ points each. For the Fold-2 checkpoint, deleting the structural input entirely moves none of the predicted probability by more than $2.1 \times 10^{-5}$. We emphasize that an inference-level measurement does not establish what the structural branch contributed during training, where it may well have acted as a useful regularizer, nor does it speak to architectures other than the one examined here. Taken with the feature-attribution results, it indicates that sequence-derived descriptors are sufficient for this task at present.

\begin{table}
\centering
\small
\caption{Feature-group attribution on the stratified 4{,}000-peptide subsample (full-feature $\mAPfive$ $80.41$; absolute values are not comparable to Table~\ref{tab:frontier}). \emph{Group-only} is a refit on that family alone, \emph{leave-one-out} is a refit without it, reported as the loss relative to the full model, and \emph{permutation} shuffles the family's columns at test time and re-infers with the same fitted model. Negative leave-one-out and permutation values denote a loss relative to the full model and positive values an improvement. All values in percentage points. The cleavage feature is a single column held out from PhysChem (Table~\ref{tab:features}).}
\label{tab:attribution}
\begin{tabular}{lrrrrr}
\toprule
Family & $n$ cols & Group-only & \% of full & Leave-one-out & Permutation \\
\midrule
ESM      & 128 & \best{79.19} & 98.5 & \best{$-0.74$} & $-19.17$ \\
CTD      & 147 & 77.30 & 96.1 & $-0.67$ & \best{$-45.19$} \\
AAC      &  20 & 77.06 & 95.8 & $-0.02$ & $-2.43$ \\
AAIndex  &  15 & 76.05 & 94.6 & $-0.21$ & $-1.70$ \\
PhysChem &  10 & 73.52 & 91.4 & $-0.07$ & $-6.45$ \\
Density  &   4 & 51.12 & 63.6 & $-0.06$ & $-0.52$ \\
Cleavage &   1 & 16.48 & 20.5 & $+0.12$ & $-0.04$ \\
Motif    &   5 & 16.26 & 20.2 & $+0.51$ & $-0.03$ \\
\midrule
All      & 330 & 80.41 & 100.0 & --- & --- \\
\bottomrule
\end{tabular}
\end{table}

\subsection{Dependence helps mainly for scarce activities}
\label{sec:dependence}
In this section we test whether the four activity labels are statistically dependent in a way that a model can use, and we measure (i) how much better a joint model scores complete multi-label outcomes than an independent-label baseline, and (ii) how much that dependence can improve ranking each activity.

Dependence is present and measurable. On held-out test data, the joint posterior gives the true four-activity label vectors a mean log-likelihood of $-0.28266$, compared with $-0.30271$ for the product of binary-relevance (independent) marginals, a gain of $+0.0200$ nats per peptide. This is dependence that can be exploited by construction.

\begin{figure}
\centering
\includegraphics[width=\linewidth]{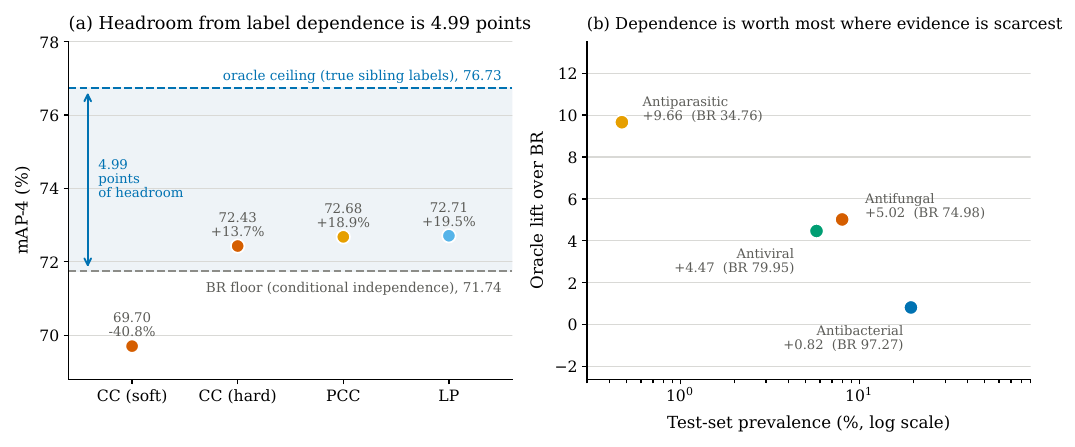}
\caption{Conditional dependence between activities is real but bounded. (a) An oracle supplied with each peptide's three true sibling activities gains 4.99 points of $\mAPfour$ over the binary-relevance floor; the best transforms recover about a fifth of that. CC recovers nothing under soft chain context and $+13.7\%$ once the context encoding matches the 0/1 encoding of the in-context labels. Percentages are the fraction of the oracle headroom recovered. (b) The oracle lift per activity, against that activity's test-set prevalence. Dependence is worth most on the rarest activities; antifungal and antiviral, whose prevalences differ by less than a factor of two, are effectively tied.}
\label{fig:dependence}
\end{figure}

For per-label ranking, however, the upside is limited. An oracle that is told each peptide's three true sibling activities reaches $\mAPfour$ $76.73$, while the binary-relevance floor is $71.74$, so the total headroom available to any method that uses sibling labels is only $4.99$ points. LP captures 19.5\% of this headroom and PCC 18.9\%. CC captures none in its default form (CC-soft), but this is an encoding mismatch rather than a failure of the factorization. If we supply hard-thresholded labels as chain context (matching the 0/1 encoding of the in-context labels), CC-hard rises from $69.70$ to $72.43$ and recovers 13.7\% of the ceiling (Figure~\ref{fig:dependence}a).

The benefit is not evenly spread across activities, and that distribution is the main practical takeaway. The oracle lift is $+0.82$ points for Antibacterial, $+4.47$ for Antiviral, $+5.02$ for Antifungal, and $+9.66$ for Antiparasitic (Figure~\ref{fig:dependence}b). Dependence tends to help most where marginal evidence is scarcest (antifungal and antiviral are effectively tied). Consistent with this, PCC's only per-label gain over BR that is individually resolvable is on Antiparasitic ($+2.91$, $[+0.81, +5.29]$), and this is what brings PCC to parity on the one label where the previous state of the art led (Table~\ref{tab:frontier}). 

One property of LP is worth noting for practitioners independent of accuracy. LP exactly respects the deterministic parent relation, i.e., it produced zero violations of the Antimicrobial OR-gate in 246{,}000 predictions across all seeds and both splits, versus violation rates of 11.3\% (PCC), 34.8\% (BR), 39.5\% (ECC), and 44.9\% (CC). A peptide called antifungal but not antimicrobial is not interpretable to a screening biologist (Table~\ref{tab:coherence}).

\begin{table}
\centering
\small
\caption{\emph{OR-gate violations} is the fraction of predictions at the honest threshold in which the predicted Antimicrobial call contradicts $\bigvee_j y_j$ over the four activities, across 246{,}000 predictions spanning all seeds and both splits.}
\label{tab:coherence}
\begin{tabular}{lrr}
\toprule
Transform & $\mAPfive$ (points) & OR-gate violations (\%) \\
\midrule
LP  & \best{77.79} & \best{0.00} \\
PCC & 77.76 & 11.33 \\
BR  & 77.01 & 34.75 \\
ECC & 76.89 & 39.51 \\
CC  & 75.27 & 44.85 \\
\bottomrule
\end{tabular}
\end{table}

\subsection{Ranking Activities from Partial Positive Evidence}
\label{sec:pu}
Curated repositories record confirmed positives and an activity a peptide was never assayed for is unlabelled rather than known-negative. The question a screening campaign actually asks is therefore not "is this peptide antifungal?" but "given what we have already confirmed, which assay should we run next?"

Scored that way the model is directly useful. The task is defined by concealing one confirmed activity and asking the model to rank it against the activities the peptide has not been confirmed for. Say, m is the number of other confirmed activities revealed to the model before it ranks. A peptide is therefore eligible only if it carries at least m+1 confirmed activities (m to reveal and one to conceal), so eligibility shrinks as m grows. 4,235 of the 16,489 test peptides carry at least one confirmed activity, 1,035 carry at least two, and 250 carry at least three. The remaining 12,254 test peptides have no confirmed activity at all, so there is nothing to conceal and no ranking to make.

At m = 0, with nothing revealed, the concealed activity is ranked first for 70.89\% of the 4,235 eligible peptides against a chance rate of 25\%, one of four candidate activities (mean reciprocal rank 0.840 on the 0–1 scale). Because nothing has been revealed, no ranker can exploit conditioning information at m = 0, and BR, LP and PCC accordingly agree to within 0.16 points; this figure measures what the sequence descriptors alone support, not what the joint model adds.

Revealing one confirmed positive raises the PCC joint posterior to $74.65\%$ $[71.95, 77.26]$ against a 33.3\% chance rate, and revealing two raises it to $89.73\%$ against 50\% (Table~\ref{tab:pu}). A permuted-label control, in which a peptide is conditioned on another peptide's confirmed activity, drops to $61.45\%$ at $m = 1$, indicating that the gain comes from the peptide-specific co-occurrence structure rather than from the generic presence of an additional input.

The joint posterior only marginally improves on binary relevance (and not consistently across $m$), consistent with the bounded ceiling in Section~\ref{sec:dependence}. The permuted control separates clearly at $m = 1$ and, on pooled positive-unlabelled average precision, remains well below the true-conditioning setting at both $m = 1$ ($92.58$ vs $68.88$) and $m = 2$ ($93.78$ vs $87.59$).

The pooled Hit@1 also conceals a strong dependence on which activity is concealed. At $m = 0$ the LP joint posterior ranks the held-out activity first for $97.77\%$ of peptides when that activity is antibacterial ($n = 2{,}157$) but $64.11\%$ when antiviral ($n = 938$), $26.59\%$ when antifungal ($n = 1{,}063$), and $3.03\%$ when antiparasitic ($n = 77$). The pooled figure is therefore dominated by the commonest target, and is the same averaging problem described in Section~\ref{sec:sota} in a different guise. The practical point is that for the three well-populated activities the ranking is far above chance in the regime where screening budgets are actually allocated.

\begin{table}
\centering
\small
\caption{Assay prioritization: rank the activities a peptide has not yet been assayed for, given $m$ already-confirmed positives. Hit@1 and pooled positive-unlabelled average precision (PU-AP) are in percentage points; mean reciprocal rank (MRR) is dimensionless and reported on the $0$--$1$ scale. Chance rises with $m$ because fewer candidate activities remain. BR cannot use the revealed positives and is the do-nothing baseline; \emph{PCC-perm} is the permuted-label control, which is undefined at $m = 0$ because nothing has been revealed.}
\label{tab:pu}
\setlength{\tabcolsep}{4.5pt}
\begin{tabular}{llrrrrr}
\toprule
Metric & $m$ ($n$ eligible) & Chance & BR & LP-joint & PCC-joint & PCC-perm \\
\midrule
Hit@1 (\%) & 0 (4{,}235) & 25.0 & 70.89 & 70.73 & 70.80 & --- \\
           & 1 (1{,}035) & 33.3 & 73.33 & 71.56 & \best{74.65} & 61.45 \\
           & 2 (250)     & 50.0 & \best{90.00} & 89.87 & 89.73 & 88.40 \\
\midrule
MRR (0--1) & 0 & --- & 0.8395 & 0.8383 & 0.8389 & --- \\
           & 1 & --- & 0.8601 & 0.8513 & \best{0.8669} & 0.8002 \\
           & 2 & --- & \best{0.9500} & 0.9493 & 0.9487 & 0.9420 \\
\midrule
PU-AP (\%) & 0 & --- & 92.75 & \best{93.17} & 92.86 & --- \\
           & 1 & --- & 91.36 & \best{92.78} & 92.58 & 68.88 \\
           & 2 & --- & 92.65 & \best{93.85} & 93.78 & 87.59 \\
\bottomrule
\end{tabular}
\begin{minipage}{0.95\textwidth}
\vspace{2pt}
\footnotesize
95\% bootstrap intervals over eligible peptides for the Hit@1 rows: at $m = 1$,
BR $[70.66, 76.01]$, LP-joint $[68.73, 74.30]$, PCC-joint $[71.95, 77.26]$,
PCC-perm $[58.42, 64.35]$. The LP permuted control gives 58.90 at $m = 1$ and 88.27
at $m = 2$.
\end{minipage}
\end{table}

\section{Discussion}
\label{sec:discussion}
Because a single antimicrobial peptide can be active against multiple pathogen classes, screening needs to predict an activity \emph{profile}, not a single label. Most work tackles this by building customized multi-output models (e.g., GNNs \citep{ruizpuentes2022}, imbalance-aware transformers \citep{pang2022transimbamp}, multi-task CNNs \citep{xu2023iampcn}, or the ESCAPE multimodal transformer \citep{escape2025}). A smaller line instead keeps a single-label learner and changes the \emph{target} via problem transformations: binary relevance, label powerset, and (deterministic/ensembled/probabilistic) classifier chains \citep{tsoumakas2007,read2011,dembczynski2010}. Although these transforms are well studied in general \citep{dembczynski2012,luaces2012,zhang2014review}, they have not been compared head-to-head on a fixed multi-activity AMP benchmark with representation and split held constant. We take that controlled route: one 330-dimensional interpretable descriptor matrix, one protocol and split, five transforms, and TabPFN as a shared backbone. Because TabPFN returns its posterior in a single forward pass over the in-context samples, each arm runs in seconds with no gradient training, which is what makes the five-way comparison practical at all.
 
Across every metric we report, label powerset (LP) transform is the most accurate (77.79 $\mAPfive$, ahead of every published method and the other four transforms). It is also the cheapest by a wide margin (one fit and one inference pass: 18\,s and 128\,s; $\times$5 faster at inference than the next-cheapest transform and $\times$32 faster than probabilistic classifier chains). Finally, it recovers the largest share of the oracle ceiling on label dependence (19.5\% vs 18.9\% for PCC and 13.7\% for the best chain), so the speed is not bought by discarding structure. LP flips the usual approach, rather than explicitly \emph{modelling} label relations, it collapses each peptide's full activity profile into a single class, so dependence is encoded as class identity. This has two practical consequences that macro metrics obscure. First, any co-occurrence pattern observed among the in-context samples is represented exactly, which is why the simplest arm leads the oracle-recovery comparison. Second, incoherent predictions are impossible because \,Antimicrobial is the logical OR of the four activities.
 
More peptide predictors now ship a predicted 3D structure alongside sequence \citep{sun2025ssfgm,yu2024toxgin,escape2025}, on the reasonable premise that membrane disruption is fundamentally structural. Our results indicate that, for the released checkpoints examined here, that modality contributes little at the point of prediction. First, a sequence-only descriptor set is already state of the art, and its signal is highly redundant: ten global physicochemical scalars recover 91.4\% of full-feature performance, and dropping any one feature family costs at most 0.74 mAP. Second, for released ESCAPE checkpoints, ablating the structural branch at inference (zero/noise/swapped maps with sequence fixed) changes ensemble macro average precision by at most 0.056, and for one checkpoint removing structure entirely changes no predicted probability by more than $2.1 \times 10^{-5}$. As distributed, these checkpoints behave as sequence-only at inference, which does not preclude the structural branch having helped during optimisation. The practical suggestion is procedural rather than critical: inference-time ablations of this kind are inexpensive, and reporting one alongside a multimodal architecture would settle the question directly, which matters most when a structure must be predicted for every screened candidate.
 
For experimental use, the more natural question is not per-label classification but \emph{which assay to run next}. In curated repositories, a missing test is recorded as 0 but is better treated as unknown. With that reading, a model that outputs a posterior over full activity profiles can prioritize hidden activities. In our setting, the concealed activity is ranked first for 70.89\% of eligible peptides (25\% by chance), improving as more positives are revealed. The caveats are that joint inference adds little beyond calibrated marginals, and performance depends strongly on which activity is concealed (97.77\% antibacterial vs 3.03\% antiparasitic). To make this a laboratory tool would require prospective ``freeze-and-test'' validation, cost-aware ranking, and an explicit positive--unlabelled treatment \citep{elkan2008,bekker2020}. The same workflow should generalize to richer label sets (e.g., species-level panels or activity--liability profiles) without changing the architecture, it needs only a feature matrix, a transform matched to the label structure, and an in-context learner.
 
Key limitations suggest clear next steps. Antiparasitic activity is too sparse (77 test positives) to support firm conclusions. Homology is handled by stratification on the predefined split rather than by holding out entire identity clusters. Finally, the prioritization analysis treats all zeros as unlabelled (so PU-AP is a lower bound) and uses a single donor assignment as a negative control rather than a permutation distribution.

\section{Conclusion}
\label{sec:conclusion}
A tabular foundation model sets a new state of the art on the ESCAPE multi-activity AMP benchmark (five-label macro AP 77.8\% vs 72.1\%) using standard sequence descriptors, and is the first to meet or exceed the best published AP on all five labels at once. It requires neither task-specific training nor structural input, and its gains over published methods are largest for remote homologues. Ablations suggest the benchmark signal is largely coarse composition, and the resulting calibrated profile probabilities enable practical next-assay ranking.

\bibliographystyle{unsrtnat}
\bibliography{references}

\clearpage
\appendix
\setcounter{section}{0}
\renewcommand{\thesection}{S\arabic{section}}
\renewcommand{\thetable}{S\arabic{table}}
\renewcommand{\thefigure}{S\arabic{figure}}
\setcounter{table}{0}
\setcounter{figure}{0}

\begin{center}
{\Large\bfseries Supplementary Material}
\end{center}

\section{The multi-label transforms, with a worked example}
\label{sec:s1}
 
\subsection{Notation}
Let $\mathbf{x} \in \mathbb{R}^{330}$ denote a peptide's descriptor vector and $\mathbf{y} \in \{0,1\}^{L}$ its label vector, with $L = 5$ in the benchmark comparison and $L = 4$ in every dependence analysis. Labels are ordered $y_1 = $ Antibacterial, $y_2 = $ Antifungal, $y_3 = $ Antiviral, $y_4 = $ Antiparasitic, $y_5 = $ Antimicrobial, and we call a complete assignment $\mathbf{y}$ a \emph{profile}. $\hat{P}$ denotes a TabPFN posterior and $\ind[\cdot]$ an indicator.
 
All five transforms wrap the same backbone on the same features and the same split. They differ only in how the five-label target is broken into problems TabPFN can solve, and therefore in what they can say about how activities co-occur.

\subsection{A worked example}
\label{sec:s1-example}
The differences are easiest to see on two labels. Consider one peptide $\mathbf{x}$ and the first two links of the chain order used throughout, Antibacterial (AB) then Antifungal (AF), and suppose the fitted models give
\begin{equation}
\hat{P}(\text{AB} = 1 \mid \mathbf{x}) = 0.6, \qquad \hat{P}(\text{AF} = 1 \mid \mathbf{x}, \text{AB} = 1) = 0.7, \qquad \hat{P}(\text{AF} = 1 \mid \mathbf{x}, \text{AB} = 0) = 0.1 .
\label{eq:s1-setup}
\end{equation}

The peptide is moderately likely to be antibacterial, and antifungal activity is far more plausible if it is. Chaining these gives a probability for each of the four profiles:
\begin{center}
\small
\begin{tabular}{lrl}
\toprule
Profile & $\hat{P}(\mathbf{y} \mid \mathbf{x})$ & \\
\midrule
$(\text{AB}=1, \text{AF}=1)$ & 0.42 & $= 0.6 \times 0.7$ \\
$(\text{AB}=1, \text{AF}=0)$ & 0.18 & $= 0.6 \times 0.3$ \\
$(\text{AB}=0, \text{AF}=1)$ & 0.04 & $= 0.4 \times 0.1$ \\
$(\text{AB}=0, \text{AF}=0)$ & 0.36 & $= 0.4 \times 0.9$ \\
\midrule
Total & 1.00 & \\
\bottomrule
\end{tabular}
\end{center}
\noindent
so the correct marginals are $\hat{P}(\text{AB}=1) = 0.42 + 0.18 = 0.60$ and $\hat{P}(\text{AF}=1) = 0.42 + 0.04 = 0.46$. Each subsection below states what each transform does with this peptide. The numbers are for illustration purpose only.
 
\subsection{Binary relevance (BR)}
$L$ independent classifiers, one per label, each seeing only $\mathbf{x}$: \begin{equation} \hat{P}_{\mathrm{BR}}(\mathbf{y} \mid \mathbf{x}) = \prod_{j=1}^{L} \hat{P}(y_j \mid \mathbf{x}) \end{equation}
A well-fitted BR recovers both marginals exactly, 0.60 and 0.46, because each classifier is estimating a marginal and nothing else. What it cannot represent is the \emph{joint}: BR assigns the doubly-active profile $0.60 \times 0.46 = 0.276$ against the 0.42 above, understating co-occurrence by 0.14. Per-label ranking metrics such as average precision are blind to this, which is why BR is competitive on $\mAPfive$ (Table~\ref{tab:frontier}) yet loses to the joint models on held-out log-likelihood (Section~\ref{sec:dependence}).
 
\subsection{Label powerset (LP)}
Each observed profile is relabelled as one class of a multiclass problem over the set $\mathcal{C}=\{11, 10, 01, 00\}$ of profiles present among the in-context samples, and marginals are recovered by summation (Equation~\ref{eq:lp}). On the example LP predicts the four-way distribution $(0.42, 0.18, 0.04, 0.36)$ directly and returns $\hat{P}(\text{AF}=1) = 0.42 + 0.04 = 0.46$, the same answer as the joint models, reached in one fit instead of a chain.
 
The cost of this is that $\mathcal{C}$ is closed. Had the profile $(\text{AB}=0, \text{AF}=1)$ never occurred among the in-context samples, LP could not assign it any probability, and its 0.04 would be redistributed over the profiles that were observed. On the benchmark this is harmless: the Antimicrobial parent is a deterministic OR, so only $2^4 = 16$ profiles are realizable and all 16 occur, well inside TabPFN's class ceiling. It is also why LP cannot violate the parent constraint, as no illegal profile is in its output space at all.

\subsection{Probabilistic classifier chains (PCC)}
At inference every profile is evaluated and the marginals recovered by summation. On the example this is the enumeration in Section~\ref{sec:s1-example}, giving $\hat{P}(\text{AF}=1) = 0.46$. PCC is the only transform here that returns a full joint posterior over profiles, which the oracle-ceiling, joint-likelihood and assay-prioritization analyses require.

Unlike LP where we only use four primary activities and compute deterministic antimicrobial activity using OR gate, PCC's output space is all $2^5$ profiles, so it can place probability mass on contradictory profiles, for example, a peptide called antifungal but not antimicrobial. This is the mechanical origin of the violation rates in Section~\ref{sec:dependence}.

\subsection{Classifier chains (CC)}
The same chain-rule factorization as PCC above, but a single path is followed rather
than every branch:
\begin{equation}
\hat{P}_{\mathrm{CC}}(\mathbf{y} \mid \mathbf{x}) = \prod_j \hat{P}(y_j \mid \mathbf{x}, y_{<j}),
\end{equation}
with $y_{<j}$ supplied as ground truth in the in-context samples and replaced at inference by the chain's own prediction. On the example, CC thresholds AB at 0.5, commits to $\text{AB} = 1$, and reports link 2's answer for that branch alone, $\hat{P}(\text{AF}=1) = 0.70$ against the correct 0.46. The 40\% of probability mass in which the peptide is \emph{not} antibacterial, and in which antifungal activity was unlikely, has been discarded. The error grows with depth, which is consistent with CC's worst per-label result falling on Antiparasitic, the fourth link (Section~\ref{sec:sota}).
 
What is appended to the next link's features also matters. Every link sees true 0/1 labels in its context. \emph{Hard} context appends $\ind[\hat{p}_{<j} > \tau]$ and so matches that encoding. \emph{Soft} context appends the raw probability $\hat{p}_{<j} \in [0,1]$, here 0.6, a value the second link never saw in context, which presents the model with an out-of-distribution input. The distinction moves CC's recovery of the oracle ceiling from $-40.8\%$ to $+13.7\%$ (Section~\ref{sec:dependence}).
 
\subsection{Ensemble classifier chains (ECC)}
CC's 0.70 above is partly an artefact of putting AB first,  under the order AF $\to$ AB the first link estimates $\hat{P}(\text{AF}=1 \mid \mathbf{x})$ with no conditioning at all, and returns something different. ECC runs eight chains under independently sampled label orders and averages each label's marginal across them, so that no single arbitrary ordering determines the result. It does not fix the committed-path problem, only spreads it over orderings. Run at TabPFN parameter $n_{\mathrm{estimators}} = 8$ to make eight chains affordable.

\section{Per-seed results}
\label{sec:s2}
\noindent Table~\ref{tab:S1-transform-benchmark} reports a per-seed benchmark of all five multi-label transforms. Every transform uses the same TabPFN v3 base classifier (\texttt{tabpfn} 8.2.0) on the same 330 sequence descriptors. We use all 65{,}870 peptides as in-context samples and evaluate on the held-out test set of 16{,}489 peptides. Three seeds ($42$, $1665$, $8914$) are reported individually together with their mean\,$\pm$\,standard deviation. \textbf{Bold} marks the best transform mean in each column.

\noindent \textbf{Average precision:} $\mAPfive$ macro-averages step-wise AP over all five labels. $\mAPfour$ excludes \emph{Antimicrobial}, which is the deterministic OR of the other four and is therefore trivially predictable. $\mAPthree$ further excludes \emph{Antiparasitic} ($n=77$ test positives), leaving the three labels for which the test set is adequately powered.

\noindent \textbf{Threshold metrics:} Max-F1 macro-averages the per-label maximum $F_1$ over thresholds chosen \emph{on the test set}. It is reported only as an optimistic upper bound. Honest $F_1$ uses per-label thresholds selected on 5-fold out-of-fold predictions over the in-context set and frozen before the test set is touched. The Max-F1 minus honest-$F_1$ gap is the cost of not knowing the threshold in advance. Subset accuracy is the exact-match rate over the full five-label vector at a fixed $0.5$ cut on every label. It understates performance on the low-prevalence labels and should be read as a conservative floor rather than a tuned operating point.

\noindent \textbf{Wall-clock:} OOF is the 5-fold out-of-fold (OOF) pass over the in-context set used to fix the honest thresholds. Fit is the final fit with all 65{,}870 in-context samples. Infer is chunked inference over the 16{,}489 test peptides. All times are seconds on a single GPU. TabPFN performs in-context learning with no gradient updates, so the fit time is dominated by ingesting the context matrix rather than by optimisation. This is why it is one to two orders of magnitude smaller than inference. Timings come from a shared machine and include contention. The CC OOF spread ($\pm$427\,s) reflects competing jobs, not a property of the transform.

\noindent ECC uses 8 chains with \texttt{n\_estimators}\,$=8$. CC and PCC use the default label order. All metric values are percentage points. The PCC subset-accuracy lead over LP is $0.01$ points, well inside the seed-to-seed spread, so it should not be read as a separation.

\begin{sidewaystable}[p]
\centering
\scriptsize
\setlength{\tabcolsep}{3pt}
\caption[Per-seed benchmark of all five multi-label transforms]{\textbf{Per-seed benchmark of all five multi-label transforms.}}
\label{tab:S1-transform-benchmark}
\resizebox{\textheight}{!}{%
\begin{tabular}{@{}llrrrrrrrrr@{}}
\toprule
& & \multicolumn{3}{c}{Average precision} & \multicolumn{3}{c}{Threshold metrics}
& \multicolumn{3}{c}{Wall-clock (s)} \\
\cmidrule(lr){3-5}\cmidrule(lr){6-8}\cmidrule(lr){9-11}
& Seed & $\mAPfive$ & $\mAPfour$ & $\mAPthree$ & Max-F1 & Honest $F_1$ & Subset acc.
& OOF & Train & Infer \\
\midrule
\multicolumn{11}{@{}l}{\textit{Binary relevance (BR)}}\\
& 42 & 76.93 & 71.64 & 84.07 & 74.17 & 72.43 & 90.56 & 1,599 & 85 & 647 \\
& 1665 & 76.90 & 71.60 & 84.07 & 73.84 & 73.04 & 90.37 & 1,595 & 84 & 640 \\
& 8914 & 77.20 & 71.98 & 84.06 & 74.09 & 72.48 & 90.46 & 1,596 & 85 & 652 \\
& \textit{mean} & 77.01\,$\pm$\,0.17 & 71.74\,$\pm$\,0.21 & 84.07\,$\pm$\,0.00 & 74.03\,$\pm$\,0.17 & 72.65\,$\pm$\,0.34 & 90.46\,$\pm$\,0.10 & 1,597\,$\pm$\,2 & 85\,$\pm$\,1 & 646\,$\pm$\,6\\
\addlinespace
\multicolumn{11}{@{}l}{\textit{Label powerset (LP)}}\\
& 42 & 78.05 & 73.03 & 84.73 & 74.26 & 73.74 & 90.75 & 321 & 18 & 128 \\
& 1665 & 77.60 & 72.47 & 84.95 & 74.19 & 73.40 & 90.82 & 319 & 18 & 128 \\
& 8914 & 77.74 & 72.64 & 84.77 & 74.18 & 73.91 & 90.70 & 328 & 18 & 128 \\
& \textit{mean} & \textbf{77.79\,$\pm$\,0.23} & \textbf{72.71\,$\pm$\,0.29} & \textbf{84.82\,$\pm$\,0.11} & \textbf{74.21\,$\pm$\,0.05} & \textbf{73.69\,$\pm$\,0.26} & 90.76\,$\pm$\,0.06 & 323\,$\pm$\,5 & 18\,$\pm$\,0 & 128\,$\pm$\,0\\
\addlinespace
\multicolumn{11}{@{}l}{\textit{Classifier chain (CC)}}\\
& 42 & 74.73 & 69.08 & 83.21 & 71.30 & 70.13 & 89.94 & 1,626 & 96 & 896 \\
& 1665 & 75.36 & 69.79 & 83.26 & 72.19 & 70.31 & 90.04 & 2,296 & 115 & 907 \\
& 8914 & 75.72 & 70.24 & 83.42 & 72.15 & 69.78 & 90.18 & 2,419 & 160 & 918 \\
& \textit{mean} & 75.27\,$\pm$\,0.50 & 69.70\,$\pm$\,0.58 & 83.30\,$\pm$\,0.11 & 71.88\,$\pm$\,0.50 & 70.07\,$\pm$\,0.27 & 90.06\,$\pm$\,0.12 & 2,114\,$\pm$\,427 & 124\,$\pm$\,33 & 907\,$\pm$\,11\\
\addlinespace
\multicolumn{11}{@{}l}{\textit{Ensemble of classifier chains (ECC)}}\\
& 42 & 76.90 & 71.66 & 83.98 & 73.31 & 71.53 & 90.39 & 8,678 & 235 & 3,805 \\
& 1665 & 76.97 & 71.76 & 83.81 & 73.27 & 73.01 & 90.11 & 8,523 & 267 & 3,779 \\
& 8914 & 76.80 & 71.56 & 84.00 & 73.44 & 72.50 & 90.41 & 8,576 & 264 & 3,788 \\
& \textit{mean} & 76.89\,$\pm$\,0.09 & 71.66\,$\pm$\,0.10 & 83.93\,$\pm$\,0.11 & 73.34\,$\pm$\,0.09 & 72.34\,$\pm$\,0.75 & 90.30\,$\pm$\,0.17 & 8,592\,$\pm$\,79 & 255\,$\pm$\,18 & 3,791\,$\pm$\,13\\
\addlinespace
\multicolumn{11}{@{}l}{\textit{Probabilistic classifier chain (PCC)}}\\
& 42 & 77.73 & 72.65 & 84.31 & 74.08 & 73.43 & 90.80 & 8,694 & 88 & 4,127 \\
& 1665 & 77.66 & 72.56 & 84.35 & 74.00 & 73.35 & 90.72 & 8,690 & 81 & 4,092 \\
& 8914 & 77.89 & 72.84 & 84.40 & 74.17 & 72.72 & 90.78 & 8,709 & 77 & 4,129 \\
& \textit{mean} & 77.76\,$\pm$\,0.12 & 72.68\,$\pm$\,0.14 & 84.35\,$\pm$\,0.04 & 74.08\,$\pm$\,0.09 & 73.16\,$\pm$\,0.39 & \textbf{90.76\,$\pm$\,0.04} & 8,698\,$\pm$\,10 & 82\,$\pm$\,6 & 4,116\,$\pm$\,21\\
\bottomrule
\end{tabular}%
}
\end{sidewaystable}

\section{Scaling with the number of in-context samples}
\label{sec:s3}
BR, LP, and CC were refitted on nested random subsets of the in-context set drawn from both folds, on the geometric ladder $n \in \{250,\, 1{,}250,\, 6{,}250,\, 31{,}250,\, 65{,}870\}$. Between the top two rungs $\mAPfive$ falls by 5.01 points for BR (76.94 to 71.93), 4.71 for LP (77.79 to 73.07), and 4.43 for CC (75.25 to 70.82). This motivated the matched-protocol experiment in Section~\ref{sec:matched}, but is not the same ablation because a random subset spans both folds, whereas the fold-matched analysis uses one fold at a time as context, and the fold-matched confound in Section~\ref{sec:matched} is correspondingly about half as large.

Table~\ref{tab:S2-scaling-sweep} uses the same held-out test set of 16{,}489 peptides. Each cell is the mean over three seeds ($42$, $1665$, $8914$). $\mAPfive$, Max-F1, and subset accuracy are reported in percentage points. Subset accuracy is the exact-match rate over the five-label vector at a fixed $0.5$ cut, while Max-F1 uses test-tuned thresholds (an optimistic upper bound).

The $65{,}870$ rows are an independent re-run of the full-data configuration and differ from Table~\ref{tab:S4-transform-benchmark} by at most $0.07$ points, bounding run-to-run variation at fixed seed.

\begin{table}[htbp]
\centering
\small
\setlength{\tabcolsep}{6pt}
\caption{Context-size scaling sweep. Macro metrics versus the number of in-context samples for BR, LP and CC.}
\label{tab:S2-scaling-sweep}
\begin{tabular}{@{}lrrrrrrrrr@{}}
\toprule
& \multicolumn{3}{c}{$\mAPfive$} & \multicolumn{3}{c}{Max-F1} & \multicolumn{3}{c}{Subset acc.} \\
\cmidrule(lr){2-4}\cmidrule(lr){5-7}\cmidrule(lr){8-10}
Context size & BR & LP & CC & BR & LP & CC & BR & LP & CC \\
\midrule
250 & 44.18 & 46.15 & 43.89 & 44.84 & 47.08 & 44.69 & 79.04 & 79.39 & 80.29 \\
1,250 & 53.93 & 54.73 & 53.93 & 53.27 & 53.55 & 53.08 & 82.26 & 82.80 & 83.25 \\
6,250 & 62.05 & 62.23 & 61.52 & 60.27 & 59.45 & 60.01 & 84.75 & 85.23 & 85.12 \\
31,250 & 71.93 & 73.07 & 70.82 & 69.17 & 69.69 & 67.59 & 88.73 & 89.15 & 88.51 \\
65,870 & 76.94 & 77.79 & 75.25 & 74.16 & 74.29 & 71.71 & 90.46 & 90.75 & 90.03 \\
\bottomrule
\end{tabular}
\end{table}

\section{Provenance of the published baselines}
\label{sec:s4}

ESCAPE authors retrained all seven prior methods from scratch on the benchmark's training folds and adapted each to emit five sigmoid outputs. Therefore no cross-dataset leakage in the published comparison. However, only two of the seven were natively designed for multi-activity prediction over these labels, and Table~\ref{tab:provenance} documents what each method's original publication actually addressed. A low row in the main-text frontier table should not be read as a verdict on the corresponding published method.

\begin{table*}[!htbp]
\centering
\footnotesize
\caption{What each published baseline's original paper addressed. \checkmark\ denotes
an activity the original method predicted; $\times$ denotes one it did not. Sample
counts are as reported in the original publication, not on the benchmark.}
\label{tab:provenance}
\setlength{\tabcolsep}{4pt}
\resizebox{\textwidth}{!}
{
\begin{tabular}{llrccccc l}
\toprule
Method & Original dataset & $n$ (original) & AB & AV & AF & AP & AM & Native task \\
\midrule
AMPs-Net \cite{ruizpuentes2022} & 19 public repositories & 23{,}967 & \checkmark & \checkmark & \checkmark & \checkmark & \checkmark & binary + 4-label \\
TransImbAMP \cite{pang2022transimbamp} & dbAMP/DRAMP/DBAASP, CD-HIT 40\% & 22{,}381 & \checkmark & \checkmark & \checkmark & \checkmark & \checkmark & binary + 7-label \\
AMP-BERT \cite{lee2023ampbert} & Veltri benchmark (APD3 + UniProt) & 3{,}556 & $\times$ & $\times$ & $\times$ & $\times$ & \checkmark & binary \\
PEP-Net \cite{han2024pepnet} & AMP and AIP benchmarks & multiple & \checkmark$^{a}$ & \checkmark$^{a}$ & \checkmark$^{a}$ & $\times$ & \checkmark & binary (separate tasks) \\
amPEPpy \cite{lawrence2021} & APD3/CAMPR3/LAMP + UniProt & 3{,}268 pos. & $\times$ & $\times$ & $\times$ & $\times$ & \checkmark & binary \\
AVP-IFT \cite{guan2024avpift} & AVPdb/dbAMP/DRAMP/DBAASP/HPIdb + Uniprot & 5{,}324 & $\times$ & \checkmark & $\times$ & $\times$ & $\times$ & antiviral specialist \\
AMPlify \cite{li2022amplify} & APD3 + DADP + UniProt & 8{,}346 & $\times$ & $\times$ & $\times$ & $\times$ & \checkmark & binary \\
ESCAPE \cite{escape2025} & 26 repositories + UniProt & 82{,}359 & \checkmark & \checkmark & \checkmark & \checkmark & \checkmark & 5-label multimodal \\
\midrule
This work & ESCAPE (unmodified) & 82{,}359 & \checkmark & \checkmark & \checkmark & \checkmark & \checkmark & 5-label, sequence only \\
\bottomrule
\end{tabular}
}
\begin{minipage}{0.95\textwidth}
\vspace{2pt}
\footnotesize
AB, antibacterial; AV, antiviral; AF, antifungal; AP, antiparasitic; AM, antimicrobial
(binary AMP vs non-AMP). $^{a}$ As separate single-activity binary datasets, not a
joint multi-label head.
\end{minipage}
\end{table*}

\end{document}